\documentclass[10pt,conference,a4paper]{IEEEtran}
\IEEEoverridecommandlockouts
\usepackage{cite}
\usepackage{amsmath,amssymb,amsfonts}
\usepackage{graphicx}
\usepackage{textcomp}
\usepackage{xcolor}
\usepackage{amsmath}
\usepackage{float}
\usepackage{multirow}
\usepackage[ruled,vlined]{algorithm2e}
\usepackage{array}
\usepackage{enumitem}
\usepackage{hyperref}
\ifCLASSOPTIONcompsoc
\usepackage[caption=false,font=normalsize,labelfon
t=sf,textfont=sf]{subfig}
\else
\usepackage[caption=false,font=footnotesize]{subfi
g}
\fi

\def\BibTeX{{\rm B\kern-.05em{\sc i\kern-.025em b}\kern-.08em
    T\kern-.1667em\lower.7ex\hbox{E}\kern-.125emX}}

\begin{document}

\setlength{\abovedisplayskip}{4pt}
\setlength{\belowdisplayskip}{4pt}
\setlength{\textfloatsep}{0pt}

\title{Motion and Region Aware Adversarial Learning for Fall Detection with Thermal Imaging
}

 \author{\parbox{18cm}{\centering
     {Vineet Mehta$^{\star}$ \qquad Abhinav Dhall$^{\dagger\star}$ \qquad Sujata Pal$^{\star}$ \qquad Shehroz S. Khan$^+$ }\\
     {\small
     $^{\star}$Indian Institute of Technology Ropar, India \qquad $^\dagger$Monash University, Australia \qquad $^+$KITE --  University Health Network, Canada}\\}
 }

\maketitle

\begin{abstract}
Automatic fall detection is a vital technology for ensuring the health and safety of people. Home-based camera systems for fall detection often put people's privacy at risk. Thermal cameras can partially or fully obfuscate facial features, thus preserving the privacy of a person. Another challenge is the less occurrence of falls in comparison to the normal activities of daily living. As fall occurs rarely, it is non-trivial to learn algorithms due to class imbalance. To handle these problems, we formulate fall detection as an anomaly detection within an adversarial framework using thermal imaging. We present a novel adversarial network that comprises of two-channel 3D convolutional autoencoders which reconstructs the thermal data and the optical flow input sequences respectively. We introduce a technique to track the region of interest, a region-based difference constraint, and a joint discriminator to compute the reconstruction error. A larger reconstruction error indicates the occurrence of a fall. The experiments on a publicly available thermal fall dataset show the superior results obtained compared to the standard baseline. 

\end{abstract}


\section{Introduction}
Automatic detection of falls is important due to the possibility of severe injury, high cost to the health system, and the psychological effect of a fall. However, due to their rarity of occurrence, traditional supervised machine learning classifiers are ill-posed for this problem \cite{khan2017review}. There are also challenges in collecting realistic fall data as it can put people’s lives in danger \cite{khan2017review}. Therefore, in many realistic settings, there may be few, or no falls data available during training. Due to these skewed data situations, we adopted fall detection as an anomaly detection problem \cite{nogas2020deepfall}. In this setting, a classifier is trained on only normal activities; during testing, both normal and fall samples are presented to the classifier.

Another challenge in video-based fall detection is preserving the person's privacy, which traditional RGB cameras cannot provide \cite{vadivelu2016thermal}. Thus, detecting falls in videos without explicitly knowing a person's identity is important for the usability of such systems in the real world. Thermal Imaging can partially or fully obfuscate a person's identity and has been used in other fall detection applications \cite{vadivelu2016thermal,nogas2018fall,nogas2020deepfall}.

Most recent works have focused on reconstruction based networks for fall detection using autoencoders \cite{nogas2020deepfall} and adversarial learning \cite{khan2020spatio}. The adversarial learning framework presents a unique opportunity to train a network to not only mimic the normal activities through a generator but also helps to discriminate it from the abnormal events through discriminator (\cite{schlegl2019f, khan2020spatio}). For video-based anomaly detection, normally, the generator is some variant of autoencoder, and the discriminator is a feed-forward neural network, both of which are trained in an adversarial manner. The most successful previous works have mostly focused on learning spatio-temporal features by using 3D Convolutional Autoencoders (3DCAE) and 3D Convolution Neural Network (3DCNN) \cite{khan2020spatio}. 

The performance of video-based fall detection may be marred by differences in background. This may become more prominent in thermal cameras where the intensities may change due to differences in heat (e.g., when a person enters the scene). Therefore, it is important to focus on the region around the person. The relative motion of the person and objects around it can also provide useful information to detect falls. 
Region and motion-based methods (\cite{li2018recurrent},\cite{carreira2017quo}) have shown superior performance in action recognition tasks. Therefore, we hypothesize that the learning spatio-temporal features utilizing region and motion awareness in video sequences would improve the detection of falls when trained in an adversarial manner. 
To this end, we propose a motion and region aware adversarial framework which consists of two separate channels optimized jointly. The first channel is input with thermal video sequence (with the extracted region of interest) and the second channel is input with corresponding optical flow. The outputs from both channels are combined to give a discriminative score for adversarial learning. We assume that joint training of thermal and optical flow channels can facilitate in the learning of both motion and region-based discriminatory features.

\section{Related Work}
There is scarce literature on detecting falls in videos in an adversarial manner with thermal cameras. We will now review studies that closely match our work.

\textbf{Fall Detection: } 
With progress in economical camera sensors, there are several works \cite{baldewijns2016bridging, sehairi2018elderly, ezatzadeh2019vifa}, which use RGB cameras for data capturing. One major limitation of RGB sensors is the lack of privacy as the identity of the subject is not preserved. To overcome this limitation, Vadivelu et al. \cite{vadivelu2016thermal} were one of the first fall detection works on thermal data. Further, Nogas et al. \cite{nogas2018fall} proposed using the thermal cameras and recurrent Convolutional AutoEncoder (CAE) for fall detection. Motivated by these works, we also use the thermal camera modality data in our experiments. The readers are also pointed to recent surveys \cite{khan2017review, ren2019research} in fall detection for more insights into different techniques proposed in the literature. Most of the recent works on fall detection using thermal and depth camera formulate it as anomaly detection.

\textbf{Anomaly Detection:}
Given the rare nature of fall events, we follow the works in abnormal event detection that is conceptually similar to our work. Many of the recent anomaly detection methods (\cite{hasan2016learning}, \cite{ravanbakhsh2017abnormal}) are based on one class classification paradigm in which the distribution of normal events are learned using autoencoders and the deviations from the learned distribution is detected as anomaly during the test time. Hasan et al. \cite{hasan2016learning} learnt the normal motion patterns in videos using hand-crafted features and CAE. Ravanbakhsh et al. \cite{ravanbakhsh2017abnormal} proposed a video to flow and vice-versa generation adversarial approach for abnormal event detection. Khan et al. \cite{khan2017detecting} proposed the use of 3DCAE for abnormal event detection applied to fall detection. Sabokrou et al.  \cite{sabokrou2018adversarially} proposed an end-to-end adversarial network that consists of a generator that reconstructs the input with added noise and the discriminator to discriminate the reconstructed output from the actual input. Further, Khan et al. \cite{khan2020spatio} extend the work of Sabokrou et al. \cite{sabokrou2018adversarially} from single image to a sequence of images for fall detection using a spatio-temporal adversarial learning framework.

Fall being a spatio-temporal change in a subject's pose, a limitation of the work is that the motion information is not explicitly added into the network. We build upon the adversarial learning work of Khan et al. \cite{khan2020spatio} and propose a two channel network,  with one channel explicitly learning the motion in the form of the optical flow while the other takes raw video frames as input. Our proposed approach can handle the situations where a person may not be present in a frame, which may reduce false positive rate.

\section{Methods}
Our proposed adversarial framework consists of two channels. The input to the first channel is a window of thermal frames and the second channel is a window of optical flow frames. Each channel consists of (i) a 3DCAE to reconstruct the input window and (ii) a 3DCNN to discriminate them from the original window of frames where both of the channels are joined by a single neuron as a joint discriminator (Fig. \ref{arch}). We trained this framework using only Activities of Daily Living (ADL) from thermal frames. We perform person tracking and extract the Region of Interest (ROI) from the thermal and the optical flow frames for motion and region-based reconstruction. 
\subsection{Advesarial Framework}
\subsubsection{3DCAE-3DCNN}
Our architecture of 3DCAE is similar to Khan et al. \cite{khan2020spatio}.  We extend their 
network by adding a channel that takes 
optical flow as input (see Table \ref{3DCAEconf}). We use 3D filters of $3\times 3$ with a temporal depth of $5$ in all layers of 3DCAE as same as Khan et al. \cite{khan2020spatio}. The operations are the same in the flow 3DCAE except for the second deconvolution layer, which uses filters of $2\times2$ with a temporal depth of $4$, to reconstruct the temporal depth of odd length.

The architecture of the 3DCNN is the same as the encoder in 3DCAE, followed by a neuron at the end with a sigmoid function to output a probability of whether a sequence of frames is original or reconstructed. Batch normalization is used in all the layers of the 3D discriminator except for the input layer. LeakyRelu activation is set in all hidden layers, with a negative slope coefficient set to 0.2 (\cite{khan2020spatio, radford2015unsupervised}).
\begin{table}[!t]
\caption{Configuration of the 3DCAE.}
\vspace{-2mm}
\label{3DCAEconf} 
\scalebox{0.96}{
\begin{tabular}{|c|c|c|}
\hline
                         & Thermal 3DCAE               & Flow 3DCAE                  \\ \hline
Input                    & (8, 64, 64, 1)                 & (7, 64, 64, 1)                 \\ \hline
\multirow{4}{*}{Encoder} & 3D Conv - (8, 64, 64, 16)   & 3D Conv - (7, 64, 64, 16)   \\ \cline{2-3} 
        & 3D Conv- (8, 32, 32, 8)     & 3D Conv- (7, 32, 32, 8)     \\ \cline{2-3} 
                         & 3D Conv - (4, 16, 16, 8)    & 3D Conv - (4, 16, 16, 8)    \\ \cline{2-3} 
                         & 3D Conv - (2, 8, 8, 8)      & 3D Conv - (2, 8, 8, 8)      \\ \hline
\multirow{4}{*}{Decoder} & 3D Deconv - (4, 16, 16, 8)  & 3D Deconv - (4, 16, 16, 8)  \\ \cline{2-3} 
                         & 3D Deconv - (8, 32, 32, 8)  & 3D Deconv - (7, 32, 32, 8)  \\ \cline{2-3} 
                         & 3D Deconv - (8, 64, 64, 16) & 3D Deconv - (7, 64, 64, 16) \\ \cline{2-3} 
                         & 3D Deconv - (8, 64, 64, 1)  & 3D Deconv - (7, 64, 64, 1)  \\ \hline
\end{tabular}}
\end{table}

\begin{figure*}[!ht]
\centering
\includegraphics[width=160mm]{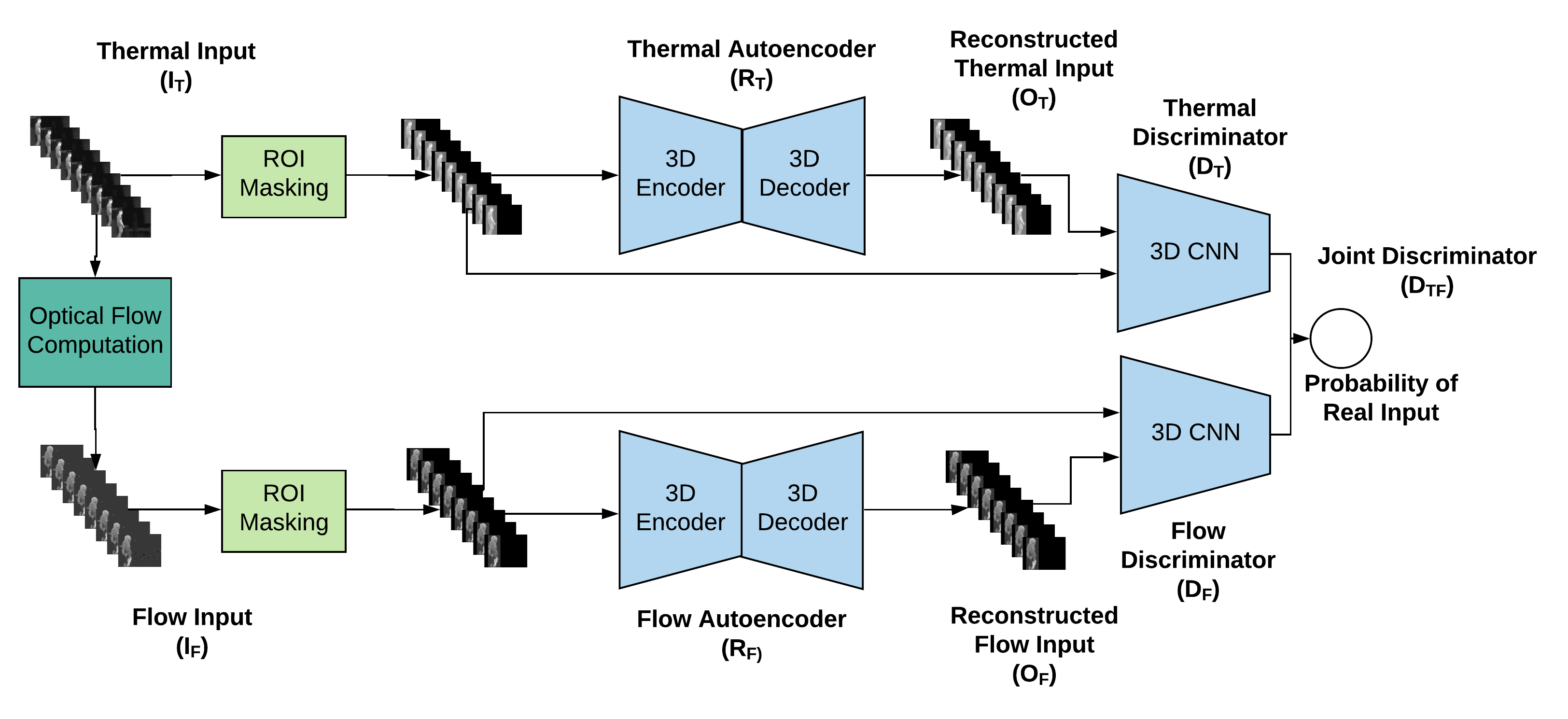} \vspace{-6mm}
\caption{The proposed adversarial network: top channel takes window of thermal frames and bottom channel takes window of the optical flow frames as input.} 
\vspace{-3mm}
\label{arch}
\end{figure*}

\subsubsection{Adversarial Learning}\label{method:adv}
In this section, we first explain the general adversarial training as described in the  work of Khan et al. \cite{khan2020spatio}. This model consists of a 3DCAE (represented as \textbf{\textit{R}}), takes the input sequence \textbf{\textit{I}} of window size \textbf{\textit{T}} and reconstructs the sequence where the output sequence is named as \textbf{\textit{O}}, which is then fed to fool 3DCNN (represented as \textbf{\textit{D}}).  \textbf{\textit{R}} and \textbf{\textit{D}} are trained using standard GAN loss described as:
\begin{equation}
L_{R+D}= E_{ I \sim p}[ \lg{D(I)}] + E_{ O \sim p}[ \lg{(1-D(O))}]\label{GANloss}
\end{equation}
We combine the adversarial loss with the Mean Squared Error (MSE) loss, which is used only for \textbf{\textit{R}} 
and defined as
\begin{equation}
L_{R} = E[ (I - O)^2 ] \label{mseloss}
\end{equation}
The total loss function to minimize 
\textbf{\textit{R}} is defined as:
\begin{equation}
L= L_{R+D} + \lambda L_{R}\label{totalloss}
\end{equation}
where $\lambda$ is a positive hyperparameter for weighted loss. The notations used for thermal and flow networks are (\textbf{\textit{I\textsubscript{T}}},\textbf{\textit{O\textsubscript{T}}},\textbf{\textit{R\textsubscript{T}}},\textbf{\textit{D\textsubscript{T}}}) and (\textbf{\textit{I\textsubscript{F}}},\textbf{\textit{O\textsubscript{F}}},\textbf{\textit{R\textsubscript{F}}},\textbf{\textit{D\textsubscript{F}}}), respectively.

\subsection{ROI Extraction} \label{ROIPipeline}
The performance of video-based fall detection methods may get impacted by background artifacts. This situation can get worse with a thermal camera because changes in the heat can alter the background and 
pixel intensity 
of frames in a video sequence \cite{nogas2020deepfall}. Therefore, we reconstruct only the region where the person is present, which is not affected much by the change in background objects and intensity. We perform person tracking using an object detector and image processing techniques to localize the person in an image.

\subsubsection{Person Detection} To the best of our knowledge, there are no pre-trained publicly available deep learning-based models for person detection, specifically for thermal images. To this end, we used the Region-based Fully Convolutional Network (R-FCN) \cite{dai2016r} trained on COCO dataset \cite{lin2014microsoft}. As the TSF dataset contains only one subject in a frame, the bounding box with the highest confidence score is selected. There are no false proposals by the detector; however, the localized bounding box is found to fluctuate in size and position, which degrades the prediction by the tracking method. 
\begin{figure}[!b]
\vspace{-2.5mm}
\centering
\subfloat{\includegraphics[width = .2\textwidth]{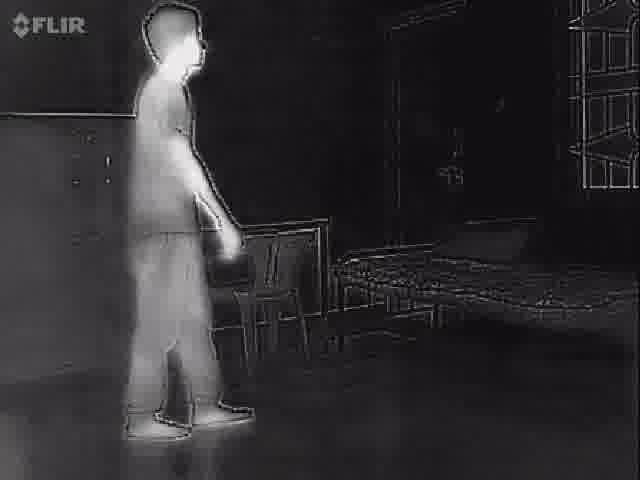}} \hspace{0.8mm} 
\subfloat{\includegraphics[width = .2\textwidth]{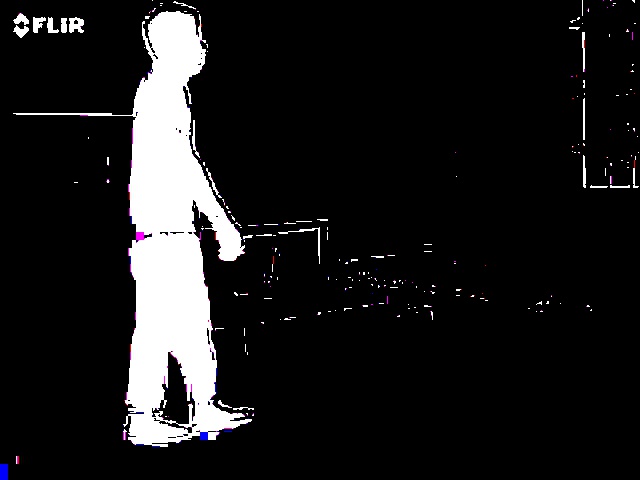}}
\\[-1ex]
\subfloat{\includegraphics[width = .2\textwidth]{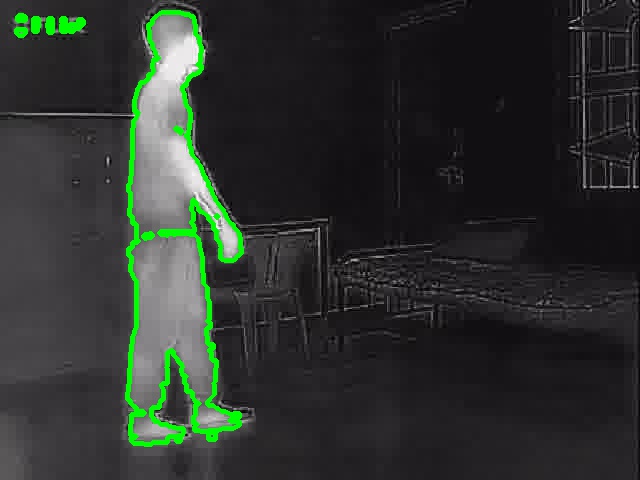}} \hspace{0.8mm}
\subfloat{\includegraphics[width = .2\textwidth]{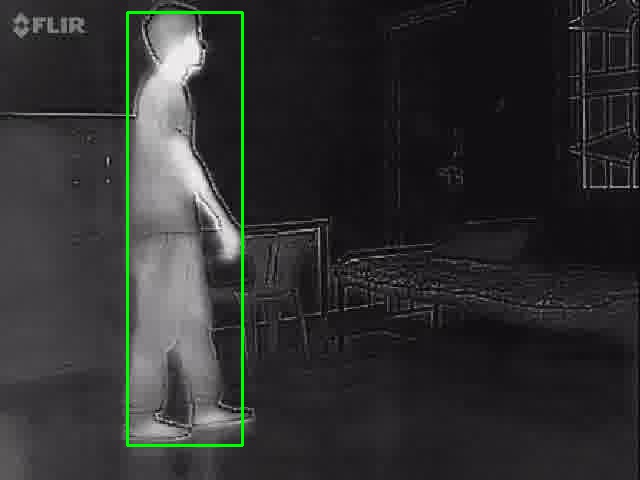}}
\vspace{-2mm}
\caption{Contour box localization process (Section \ref{contourbox section}).} 
\label{contourbox}
\end{figure}

\subsubsection{Contour Box Localization} \label{contourbox section}
 In thermal images, a person may appear brighter than the background due to differences in heat emitted by as a person and object. Therefore, Otsu thresholding\cite{otsu1979threshold} is applied to the thermal image to separate dark background, as shown in Fig. \ref{contourbox}. The thresholded image may still contain bright background objects. We find the contours \cite{suzuki1985topological} on the thresholded image after applying morphological operations and select the biggest contour on the basis of the inside area. The smallest box containing that contour blob is chosen as a candidate for the person bounding box.

\subsubsection{Tracking}
We apply Kalman filtering on the top-left and right-bottom coordinates of the bounding box with the constant velocity assumption. The tracker is initialized with the person detector and predicts the bounding box for the next frame. We compare the predicted box with the person detection bounding box (if detected in the next frame) to check if the tracker drifts. We use a counter (age) to track the number of continuous tracker predictions without detection. In the case of no detection, the tracker's age is increased, and when the age of the tracker exceeds a limit of 20, the tracker is stopped. The Intersection of Union (IoU) is used in many tracking methods to match the bounding boxes. However, IoU is small when the size of one box is large compared to the other box, which could be possible due to the bad localization by the detector. Therefore, we also used other criteria such as the ratio of area and checking for a subset. At a particular instant, there could be at most three possible candidates for the person localization -- Detect, Contour, and Track box. Detect box confirms the presence of the person, but it does not fit the person in most cases. Therefore, the contour box and track box are used to improve the overall tracking. Algorithm \ref{alg:trackingalgo} describes the whole tracking method. 

\begin{algorithm}[!ht]
\caption{Tracking Algorithm}
\label{alg:trackingalgo}
    \SetKwInOut{Input}{Input}
    \SetKwInOut{Output}{Output}
    
    \Input{Frame}
    \Output{FinalBox}
    \SetAlgoLined
    FinalBox=None\;
    DetectBox=Detector.GetLocalization(Frame)\;
    \eIf{DetectBox is not None}{
        ContourBox=GetBiggestContourBox(Frame)\;
        \If{ Tracker is not None}{
            TrackBox=Tracker.GetCurrentBox()\;
            \eIf{TrackBox matches with DetectBox}{
                DetectBox=BoxSelection(DetectBox,TrackBox)\;
            }{
                Tracker=None\;
            }
        }
        \If{ContourBox matches with DetectBox}
        {
            DetectBox=BoxSelection(DetectBox,ContourBox)\;
        }
        FinalBox=DetectBox\;
        \eIf{Tracker is not None}{
            Tracker.KalmanFilter(DetectBox)\;
            Tracker.Losses=0\;
        }{
            Tracker=InitializeTracker(DetectBox)\;
        }
    }{
        \If{Tracker is not None}{
            TrackBox=Tracker.Predict()\;
			ContourBox=GetBiggestContourBox(Frame)\;
			\eIf{ContourBox matches with TrackBox}{
                Tracker.Losses+=0.5\;
                FinalBox=ContourBox\;
                Tracker.KalmanFilter(ContourBox)\;
			}{
    			Tracker.CurrentBox=TrackBox\;
    			FinalBox=TrackBox\;
    			Tracker.Losses+=1\;
			}
        }
    }
    \If{Tracker is not None and Tracker.Losses greater than MaxAge}{
        Tracker=None\;
        FinalBox=None\;
    }
    Return FinalBox\;
\end{algorithm} 

\subsubsection{Region based reconstruction}
We remove the frames in which a person is not localized post tracking, and rest of the frames are masked by their corresponding bounding box (named as ROI mask). For region-based reconstruction, the 3DCAE is fed with the window of masked frames and a region-based reconstruction loss \textit{L\textsubscript{ROI}} (instead of \textit{L\textsubscript{R}}) is used: 
 \begin{equation}
L_{ROI} = E_{ROI}[ ( ROI(I) - ROI(O) )^2 ] \label{roiloss}
\end{equation}

where ROI(X) represents the masking of frames in window X with the corresponding ROI masks, and the expectation is taken over pixels inside the ROI.

\subsection{Motion Constraint and Reconstruction}
Beside the appearance based constraint on \textbf{\textit{R}}, we incorporate
motion in the fall detection system in two ways:

\subsubsection{Difference Constraint}
Mathieu et al. \cite{mathieu2015deep} compute the difference and gradient-based losses for future frame prediction, which increase the sharpness of the predicted frame.
We adapted a similar technique to add an additional loss term in the thermal 3DCAE, which is based on the MSE of the difference frames of \textbf{\textit{I}} and \textbf{\textit{O}}. A difference frame is a residual map computed by subtracting two consecutive frames. We mask the difference frames by their respective ROI, which is the union of the ROIs of the two frames used to compute the difference frame. Further, the difference loss is defined as:
\begin{equation}
L_{Diff} = E_{ROI}[ ( ROI(DF(I)) - ROI(DF(O)) )^2 ] \label{diffroiloss}
\end{equation}
where DF(X) represents the difference frames for the window X.
Therefore, the final loss for $R$ in \eqref{totalloss} with \textit{L\textsubscript{ROI}} and \textit{L\textsubscript{Diff}} is defined as:
\begin{equation}
L= L_{R+D} + \lambda_{S}L_{ROI} + \lambda_{D}L_{Diff}\label{diffTotalLoss}
\end{equation}
where $\lambda_{S}$ and $\lambda_{D}$ are positive constants for weighted loss.
   
\subsubsection{Optical Flow Reconstruction}
Liu et al. \cite{liu2018future} use constraint on the optical flow of the ground truth and reconstructed images for future frame prediction. They used CNN based model for flow estimation, which can facilitate backpropagation for optical flow loss, but these models are trained on RGB images and may generate noise due to temperature changes in a thermal video. Therefore, we train a spatio-temporal adversarial network (\textbf{\textit{R\textsubscript{F}}}, \textbf{\textit{D\textsubscript{F}}}) for flow reconstruction which takes into input a window of optical flow frames. We compute the dense optical flow frames for two consecutive frames \cite{farneback2003two}. We stack the flow in the x, y direction, and the magnitude to form a 3-dimensional image (similar to \cite{Gkioxari_2015_CVPR}).

The flow images are masked with their ROI to remove noise due to temperature variations.
As defined earlier for the difference frame, the ROI for flow image is the union of ROI of the two thermal frames used to compute the optical flow.
\subsection{Thermal and Optical Flow Fusion}\label{method:fusion}
We propose a joint adversarial network for thermal and optical flow reconstruction, which consists of two 3DCAE and a single joint 
discriminator. In the joint discriminator network \textbf{\textit{D\textsubscript{TF}}}, the individual 3DCNN networks - \textbf{\textit{D\textsubscript{T}}} and \textbf{\textit{D\textsubscript{F}}} are used where the two individual sigmoid neurons are replaced by a single sigmoid neuron (see Figure \ref{arch}). For region based reconstruction, the joint loss function is defined as:
\begin{equation}
L= L_{R+D} + \lambda_{T}L_{T\_ ROI} + \lambda_{F}L_{F\_ ROI}
\label{fusionLoss}
\end{equation}
where \textit{L\textsubscript{T\textunderscore ROI}} and \textit{L\textsubscript{F\textunderscore ROI}} are region-based reconstruction loss for thermal and optical flow, respectively. The hyperparameters  $\lambda_{T}$ and $\lambda_{F}$ are their corresponding positive constants for weighted loss.
The joint loss function with region based reconstruction and difference constraint can be written as:-
\begin{equation}
L= L_{R+D} + \lambda_{T\_ S}L_{T\_ ROI} + \lambda_{T\_ D}L_{T\_ Diff} + \lambda_{F}L_{F\_ ROI}
\label{fusionDiffLoss} 
\end{equation}
where L\textsubscript{T\textunderscore Diff} (Eq. \eqref{diffroiloss}) and L\textsubscript{T\textunderscore ROI} (Eq. \eqref{roiloss}) are difference constraint and region based reconstruction loss. 

\section{Detecting Unseen Falls}
\begin{figure}[!t]
  \centering 
    \includegraphics[width=.36\textwidth]{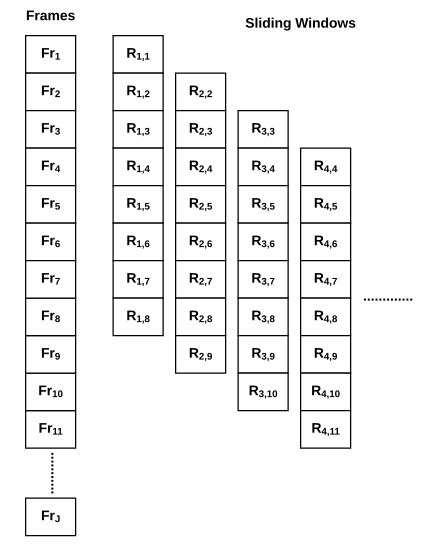} 
    \vspace{-3mm}
   \caption{Temporal sliding window method- showing error (R\textsubscript{i,j}) per frame (Fr\textsubscript{j}) with T = 8 (Figure Source: \cite{khan2020spatio}).}
   \label{score} 
\end{figure}

The strategy to detect unseen falls is shown in Figure \ref{score} (derived from \cite{nogas2020deepfall}). All the frames in the video, Fr\textsubscript{i}, are broken down into windows of frames of length, \textbf{\textit{T}} = $8$, using the sliding window method with stride=$1$. For the $i^{th}$ window $I_{i}$, the 3DCAE gives the output as a reconstruction of this window, $O_{i}$. The reconstruction error (R\textsubscript{i,j}) between the $j^{th}$ frame of $I_{i}$ and $O_{i}$ can be calculated as (similar to Eq. \eqref{mseloss})
\begin{equation}
R_{i,j} = E[ (I_{i,j} - O_{i,j})^2 ] \label{RE}
\end{equation}
For the region based reconstruction models, it can be defined (similar to Eq. \eqref{roiloss}) as follows:
\begin{equation}
R_{i,j} = E_{ROI}[ (ROI(I_{i,j}) - ROI(O_{i,j}))^2 ] \label{ROIRE}
\end{equation}
There are two ways to detect unseen falls, 
at the frame level, or 
at the window level, which are described below.

\subsection{Frame Level Anomaly Score}
In the frame level anomaly detection, the reconstruction error (R\textsubscript{i,j}) (obtained from the 3DCAE) is computed for every $j^{th}$ frame across different windows. The average (C\textsuperscript{j}\textsubscript{$\mu$}) and standard deviation (C\textsuperscript{j}\textsubscript{$\sigma$}) of the R\textsubscript{i,j} across different windows are used as an anomaly score for the $j^{th}$ frame as follows \cite{nogas2020deepfall}:
\begin{equation}
\begin{aligned}
C_{\mu}^{j}=\begin{cases}\frac{1}{j}\sum_{i=1}^j R_{i,j} & j < T\\\frac{1}{T}\sum_{i=j-T+1}^j R_{i,j} & j \geq T\end{cases}
\\
C_{\sigma}^{j}=\begin{cases}\sqrt{\frac{1}{j}\sum_{i=1}^j (R_{i,j}-C_{\mu}^{j})^2} & j < T\\\sqrt{\frac{1}{T}\sum_{i=j-T+1}^j (R_{i,j}-C_{\mu}^{j})^2} & j \geq T\end{cases}
\end{aligned}
\label{framescore}
\end{equation}

\subsection{Window Level Anomaly Score} \label{slidingwindow}
For window level anomaly detection, 
the anomaly score is calculated for the entire window. For a particular window W\textsubscript{i} of length \textbf{\textit{T}}, the mean of reconstruction error of all the \textbf{\textit{T}} frames (W\textsuperscript{i}\textsubscript{$\mu$}) and their standard deviation (W\textsuperscript{i}\textsubscript{$\mu$}) are used as an anomaly score \cite{khan2020spatio}:
\begin{equation}
\begin{aligned}
W_{\mu}^{i}=\frac{1}{T}\sum_{j=i}^{T+i-1} R_{i,j}\\
W_{\sigma}^{i}=\sqrt{\frac{1}{T}\sum_{j=i}^{T+i-1} (R_{i,j}-W_{\mu}^{i})^2}
\end{aligned}
\label{windowscore}
\end{equation}
While using difference constraint in thermal reconstruction models, we also compute another window level anomaly score using the reconstruction error of difference frames (Eq. \eqref{diffroiloss}) by taking the mean and standard deviation of reconstruction error over \textbf{\textit{T}}-$1$ frames (similar to Eq. \eqref{windowscore} with window length \textbf{\textit{T}}-$1$). Similarly, we compute the window level anomaly scores for the optical flow reconstruction models using the reconstruction error of the optical flow frames.  

For region-based reconstruction models with difference constraint, we name the anomaly scores computed using thermal frames as \textbf{ROI-score} and anomaly scores computed using difference frames as \textbf{Diff-score}. For fusion models, we use the terms \textbf{Thermal ROI-score}, \textbf{Flow ROI-score} and \textbf{Thermal Diff-score} for comparisons of the different anomaly scores calculated for the same model. We define tolerance ($\alpha$) as the number of fall frames required 
in a window to set the ground-truth label of the entire window as a fall. We varied $\alpha$ from $1$ to \textbf{\textit{T}} 
to understand its impact on the results.

\section{Experiments}
\begin{figure}[t]
\centering
\subfloat{\includegraphics[width = .22\textwidth,height=2.3cm]{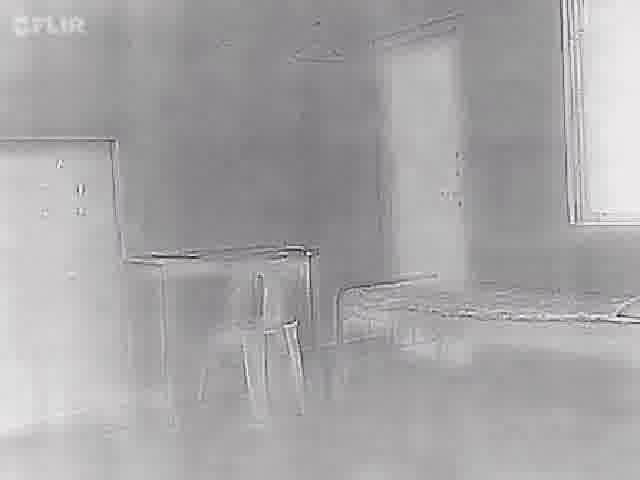}} \hspace{0.8mm}
\subfloat{\includegraphics[width = .22\textwidth,height=2.3cm]{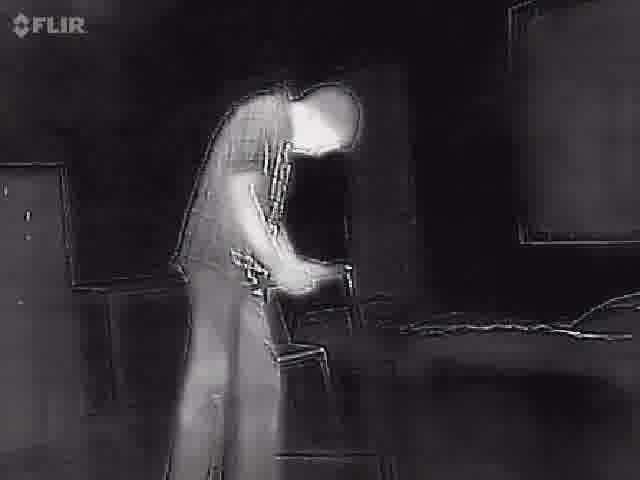}}
\\[-1ex]
\subfloat{\includegraphics[width = .22\textwidth,height=2.3cm]{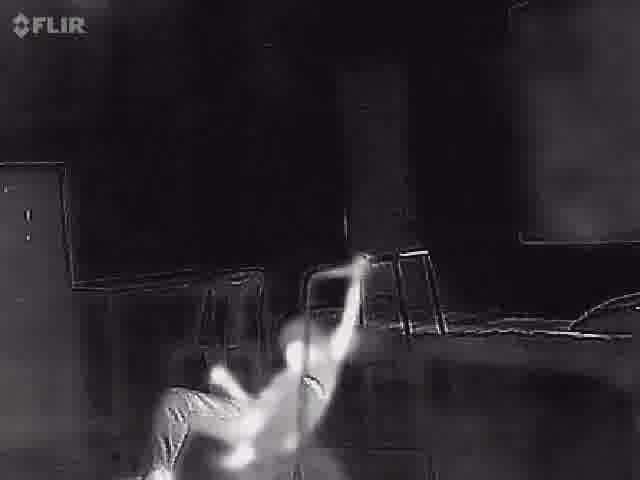}}\hspace{0.8mm}
\subfloat{\includegraphics[width = .22\textwidth,height=2.3cm]{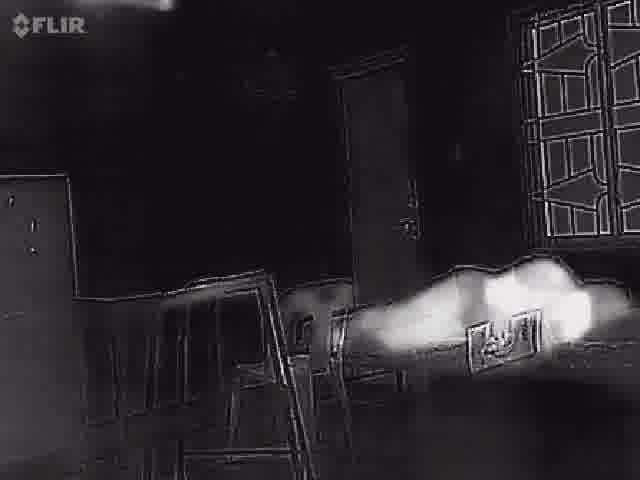}}
\vspace{-2mm}
\caption{TSF dataset \cite{vadivelu2016thermal} samples: top-left shows empty room frame; bottom-left shows the fall while during walking and the right column shows ADL frames-walking and lying on bed respectively.} 
\label{dataset}
\end{figure}

\textbf{Dataset:}
We use the publicly available TSF dataset \cite{vadivelu2016thermal} containing 44 thermal videos of resolution  640x480. There are 9 videos with normal ADL and 35 videos containing falls and other normal activities. The ADL frames include different scenarios such as an empty room, a person entering a room, sitting on a chair, or lying in bed, whereas the fall frames include a person falling from the chair, bed, or falling while walking. Some of the ADL and fall frames are shown in Fig. \ref{dataset}. ADL videos contain a total of 22,116 frames.

\begin{figure*}[!htbp]
\centering
    \subfloat[3DCAE input]{\includegraphics[width = .18\textwidth]{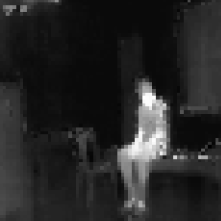}}
    \hskip .5em
    \subfloat[3DCAE output]{\includegraphics[width = .18\textwidth]{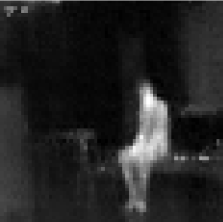}}
    \hskip.5em
    \subfloat[ROI-3DCAE input]{\includegraphics[width = .18\textwidth]{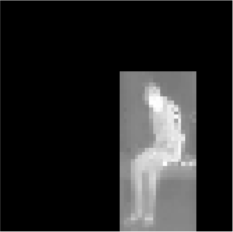}}
    \hskip .5em
    \subfloat[ROI-3DCAE output]{\includegraphics[width = .18\textwidth]{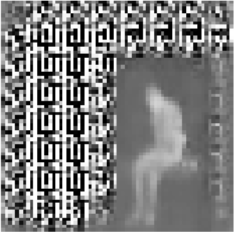}}
    \hskip .5em
    \subfloat[ROI-3DCAE masked output]{\includegraphics[width = .18\textwidth]{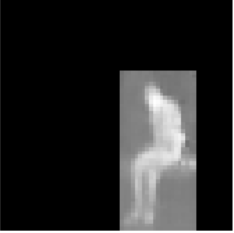}}
    \\[-2ex]
    \subfloat[3DCAE flow input]{\includegraphics[width = .18\textwidth]{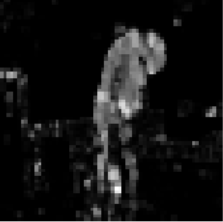}}
    \hskip .5em
    \subfloat[3DCAE flow output]{\includegraphics[width = .18\textwidth]{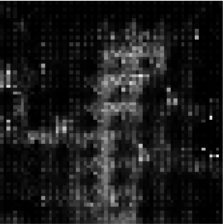}}
    \hskip.5em
    \subfloat[ROI-3DCAE flow input]{\includegraphics[width = .18\textwidth]{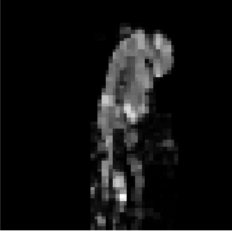}}
    \hskip .5em
    \subfloat[ROI-3DCAE flow output]{\includegraphics[width = .18\textwidth]{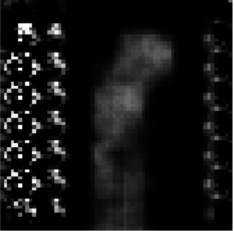}}
    \hskip .5em
    \subfloat[ROI-3DCAE masked flow output]{\includegraphics[width = .18\textwidth]{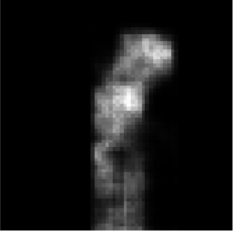}}
 \vspace{-2mm}
\caption{Qualitative analysis- The middle frame of the input and output window reconstructed by different models: the top row images [(a)-(e)] and bottom row images [(f)-(j)] are the inputs and corresponding outputs of the thermal data channel and flow magnitude channels, respectively.}\label{outputPipeline} 
\vspace{-3mm}
\end{figure*}

\textbf{Data Processing:}
We perform person tracking on thermal videos; when a person is not detected continuously for $10$ frames, we break the video and create sub videos, resulting in $22$ sub videos, which further contain 12,454 frames from ADL videos for training. For ROI computation, the detection threshold is empirically set to $0.3$. We perform the sliding window method on sub videos with stride=1 to create windows of length \textbf{\textit{T}}=8, which creates a total of 12,300 windows from ADL sub videos.
We normalize frame in the range [-1, 1] \cite{radford2015unsupervised} followed by resizing to 64x64. 

\textbf{ROI processing: } For the input to region-based networks, we perform normalization on the pixels inside the ROI and mask it by ROI mask. We set the pixels outside the ROI to $-1$. Finally, an ROI masked frame contains pixel value in the range $[-1, 1]$ inside the ROI and $-1$ outside the ROI.

\textbf{Optical flow image pre-processing}: Each channel of a flow image is re-scaled in the range of $[-1, 1]$ by min-max normalization. For region-based reconstruction in flow images, we perform ROI masking as described earlier for the normalized thermal image, where the ROI of the optical flow frame is the union of the corresponding consecutive frames.

\subsection{Network Implementations}\label{exp:Network Implementations}
We train two adversarial models, one each for for thermal window (\textbf{Thermal-3DCAE}) and optical flow window (\textbf{Flow-3DCAE}) reconstruction.
For region-based reconstruction, we train two adversarial models on thermal data, one with ROI masking and ROI loss (Eq. \eqref{roiloss}) described earlier and the other one with the addition of difference constraint in the region based reconstruction (named as \textbf{Thermal-ROI-3DCAE} and \textbf{Thermal-Diff-ROI-3DCAE}) 
We train one adversarial model for optical flow with region-based reconstruction, named as \textbf{Flow-ROI-3DCAE}.
For the fusion networks, we train two models: one without the difference constraint (\textbf{Fusion-ROI-3DCAE}) and one with it (\textbf{Fusion-Diff-ROI-3DCAE}).

We use the SGD optimizer with the learning rate as $0.0002$ for the 3DCNN discriminator and adadelta optimizer for the 3DCAE in all the adversarial models. All the models are trained for 300 epochs. The hyperparameters (\textbf{$\lambda$}'s) used for the weighted loss (Eq-\eqref{totalloss}, \eqref{diffTotalLoss} and \eqref{fusionLoss}) are varied between three values [0.1, 1 and 10]. We found that the large value of these hyperparameters led to mode collapse. The best hyperparameter setting for \textbf{Thermal-Diff-ROI-3DCAE} and \textbf{Fusion-Diff-ROI-3DCAE} have all the constants equal to 1, whereas the hyperparameter setting for the rest of the models has all the constants equal to 0.1. The full code of our implementation is available at \textbf{\url{https://github.com/ivineetm007/Fall-detection}}. 

\begin{figure}[!b]
\centering 
\includegraphics[width=0.45\textwidth]{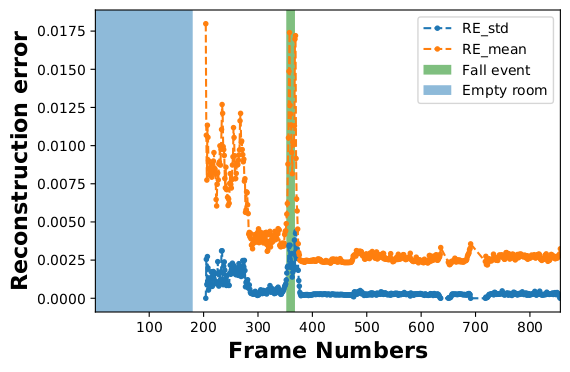} \vspace{-5mm}
\caption{Frame level anomaly score for a fall video.} 
\label{fig:visual}
\end{figure}

\begin{figure*}[!ht]
\centering 
\includegraphics[width=\textwidth]{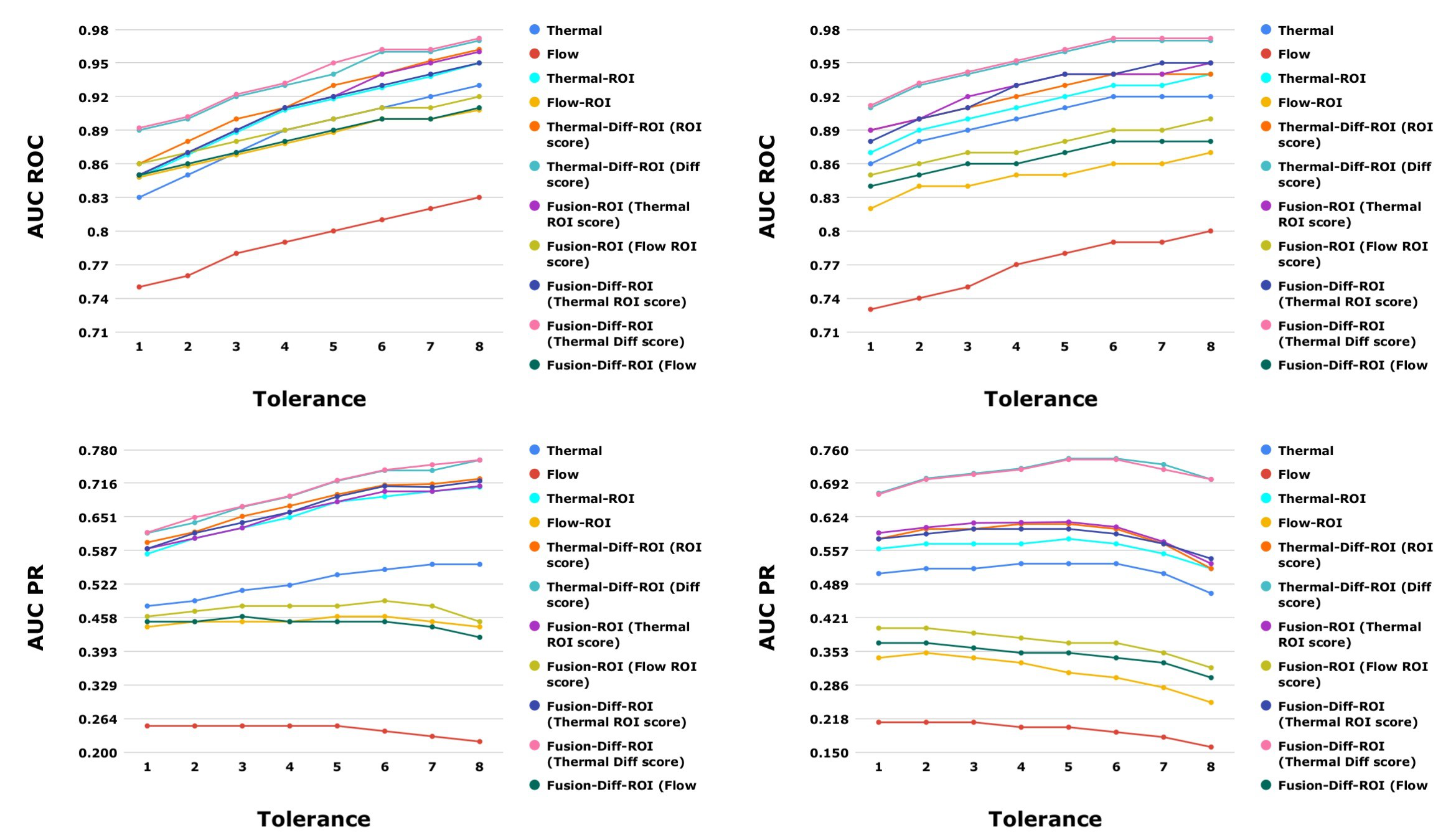} \vspace{-9mm}
\caption{Plots of AUC values of ROC and PR curve computed using window level anomaly scores with the variation in tolerance- \textbf{W\textsubscript{$\mu$}}(Left) and \textbf{W\textsubscript{$\sigma$}}(Right). }\label{fig:windowscores} 
\vspace{-3mm}
\end{figure*}
\subsection{Evaluation Metrics}
For assessing the performance of detecting falls as anomaly, the Area Under Curve (AUC) of Receiver Operating Characteristics (ROC) and Precision-Recall (PR) curve is used. The latter is used to specifically focus on the detection of minority `fall' class. We compute the anomaly scores at the frame and window level, which consists of two types - mean and standard deviation anomaly scores. We calculate and compare the AUC of ROC and PR curve using all these scores.

\subsection{Ablation Studies}\label{sec:ablation studies}
\subsubsection{3DCAE}
The thermal input and the reconstructed output by \textbf{Thermal-3DCAE} can be seen in Fig. \ref{outputPipeline}. The basic techniques to utilize the ROI in the deep learning models are resizing and ROI pooling. We also train two different \textbf{Thermal-3DCAE} models by changing the thermal input (1) resizing input ROI to 64x64, and (2) ROI pooling to 64x64 dimension. We observe that these techniques increase the false-positives, and the results for them are not reported. We argue that the resizing the ROI leads to geometric distortions and introduces false motion on the borders even if the subject is not moving.

\textbf{Optical Flow:} As described earlier, optical flow images contain noise due to temperature variation 
due to which the reconstruction quality is also noisy (Fig. \ref{outputPipeline}(g)).

\subsubsection{ROI-3DCAE}
The output of \textbf{Thermal-ROI-3DCAE} and \textbf{Flow-ROI-3DCAE} 
is shown in Fig. \ref{outputPipeline}. We observe that the region-based method improves the reconstruction quality in the ROI region as the model learns to reconstruct only the ROI (see Fig. \ref{outputPipeline} (c), (d), and (e)). Similar behaviour is observed for \textbf{Flow-ROI-3DCAE} (see Fig. \ref{outputPipeline} (h), (i) and (j)).

\subsubsection{Difference constraint}\label{sec:exp-difference constraint}
To understand the impact of difference constraint, we compare the results computed at window level (see Fig. \ref{fig:windowscores}) for the AUC of ROC and PR. On the comparison between the Diff score and the ROI score of the \textbf{Thermal-Diff-ROI-3DCAE}, we found that the results using the Diff score are better than the ROI score in all the plots, which suggests that the use of Diff score is more suitable for window level analysis. 
We also compare the results of \textbf{Thermal-ROI-3DCAE} and Diff score of \textbf{Thermal-Diff-ROI-3DCAE}, we found that the addition of constraint increase the AUC for both ROC and PR curve.  Similar behaviour was observed on the comparison of \textbf{Fusion-ROI-3DCAE} and \textbf{Fusion-Diff-ROI-3DCAE}. This suggests that the difference constraint makes the model more discriminative in temporal direction and increases the overall 
performance.

\renewcommand{\tabcolsep}{6pt}
\begin{table}[!b]
\centering
\caption{ AUC of ROC and PR based frame level anomaly comparison.} 
\vspace{-2mm}
\label{tab:framescores}
\scalebox{0.93}{
\begin{tabular}{|l|r|r|r|r|}
\hline
\textbf{Method}       & \multicolumn{2}{c|}{\textbf{ROC}} & \multicolumn{2}{c|}{\textbf{PR}}                                  \\ \hline
                      & \multicolumn{1}{c|}{\textbf{C\textsubscript{$\mu$}}} & \multicolumn{1}{c|}{\textbf{C\textsubscript{$\sigma$}}}& \multicolumn{1}{c|}{\textbf{C\textsubscript{$\mu$}}} & \multicolumn{1}{c|}{\textbf{C\textsubscript{$\sigma$}}} \\ \hline
Thermal-3DCAE           & 0.88                               & 0.90 & 0.47                               & 0.48                              \\ \hline
Thermal-ROI-3DCAE           & 0.89                               & 0.92 & 0.55                             & 0.57                              \\ \hline
Thermal-Diff-ROI-3DCAE (ROI score)           & 0.90                               & 0.92 & 0.57                             & 0.56                              \\ \hline
Fusion-ROI-3DCAE (Thermal ROI score)           & 0.90                               & 0.93 & 0.56                             & \textbf{0.58}                              \\ \hline
Fusion-Diff-ROI-3DCAE (Thermal ROI score)           & \textbf{0.90}                               & \textbf{0.93} & \textbf{0.57}                             & 0.57                              \\ \hline
\end{tabular}}
\end{table}

\subsubsection{Fusion models}
To understand the effect of fusion, we first compare the results of \textbf{Fusion-ROI-3DCAE} with the ROI based models at frame level as shown in Table \ref{tab:framescores}; we found that there are minor improvements in frame-level results. We compare the window level results ( Fig. \ref{fig:windowscores}) of \textbf{Thermal-ROI-3DCAE} and \textbf{Flow-ROI-3DCAE} with Thermal-ROI score and Flow-ROI score of \textbf{Fusion-ROI-3DCAE} respectively, we observe that there is a small increment in the results by ROI score of thermal and a substantial increment in the results by ROI score of flow, which indicates that joint learning improved the learning of flow reconstructor.

\textbf{Qualitative Analysis}- We use the frame based anomaly scores to visualize the performance of the proposed method (see Fig. \ref{fig:visual}). Although our model is able to detect fall events, we observed a false peak when the person enters the room and our tracking method misses some frames when person enters.

\section{Results}\label{sec:results}
The AUC values of ROC and PR curves computed using frame level anomaly scores are shown in Table \ref{tab:framescores}. For window level results, we plot 
the AUC values with the variation of tolerance ($\alpha$) from $1$ to $8$ 
(see Fig. \ref{fig:windowscores}). The frame level and window level results can be summarized as:

\begin{enumerate}[leftmargin=*]
  \item There is a improvement in the AUC of the ROC and PR curve by region-based construction, which confirms the importance of region awareness. 
  \item Addition of difference constraint increases the AUC values using window level scores, which indicates its importance for the learning of spatio-temporal autoencoders.
  \item Fusion models leads to an increase in the performance.
\end{enumerate}

\textbf{Comparison with the existing methods: }In the previous works using DSTCAE-C3D \cite{nogas2020deepfall}, Conv-LSTM AE \cite{nogas2018fall} and 3DCAE-3DCNN \cite{khan2020spatio}, AUC values of ROC computed using frame-level anomaly score are reported (see Table \ref{comparisontable} `All frames' column). The previous methods do not perform person tracking in the video, due to which the number of frames used for training and testing is different. Therefore, we train and test these methods using the frames on which a person is localized by our tracking method. The comparison of these methods with the proposed model on tracked frames only is shown in Table \ref{comparisontable} and summarized as:
 \begin{enumerate}[leftmargin=*]
  \item AUC values of previous methods decrease when only tracked frames are used. Furthermore, the empty frames in train and test set videos are similar. Low reconstruction error on these frames may give high AUC values in previous methods (Table \ref{comparisontable}-`All frames') during testing.
 \item The proposed methods achieves similar and better AUC of ROC than previous methods and higher AUC of PR against all previous methods. The method focuses on region in the frame where a person is present; therefore, it can facilitate learning of background agnostic models.
 \end{enumerate} 
 \renewcommand{\tabcolsep}{6pt}

\begin{table}[!t]
\centering
\caption{Comparison with the previous methods based on AUC of ROC and PR curve calculated on frame level anomaly scores.}
\vspace{-2mm}
\label{comparisontable}
\scalebox{0.95}{
\begin{tabular}{|l|r|r|r|r|r|r|}
\hline
\textbf{Method}       & \multicolumn{2}{c|}{\textbf{All frames }} & \multicolumn{4}{c|}{\textbf{Tracked frames }}                                  \\ \hline
\textbf{}       & \multicolumn{2}{c|}{\textbf{ROC}} & \multicolumn{2}{c|}{\textbf{ROC}} & \multicolumn{2}{c|}{\textbf{PR}}                                  \\ \hline
                      & \multicolumn{1}{c|}{\textbf{C\textsubscript{$\mu$}}} & \multicolumn{1}{c|}{\textbf{C\textsubscript{$\sigma$}}}
                      & \multicolumn{1}{c|}{\textbf{C\textsubscript{$\mu$}}} & \multicolumn{1}{c|}{\textbf{C\textsubscript{$\sigma$}}} & \multicolumn{1}{c|}{\textbf{C\textsubscript{$\mu$}}} & \multicolumn{1}{c|}{\textbf{C\textsubscript{$\sigma$}}}\\ \hline
Conv-LSTM AE\cite{nogas2018fall}           & 0.76                               & 0.83          & 0.63                               & 0.73 & 0.26                               & 0.37                    \\ \hline
DSTCAE-C3D\cite{nogas2020deepfall}             & 0.93                               & 0.97          & 0.85                               & 0.90 & 0.46                               & 0.53                   \\ \hline
3DCAE-3DCNN\cite{khan2020spatio}           & 0.95                               & 0.95                  & \textbf{0.90}                               & 0.88           & 0.47                               & 0.48 \\ \hline
Fusion-Diff-ROI-3DCAE & ---                               & --- & \textbf{0.90}                               & \textbf{0.93} & \textbf{0.57}                               & \textbf{0.57}    \\    \hline                        
\end{tabular}}
\end{table}

\section{Conclusion and Future Work}
Fall detection is a non-trivial problem due to large imbalance in data; thus, we formulate it as an anomaly detection problem. In this problem, we trained the model on only normal ADL and predicts whether the test sample is normal ADL or a fall. Building upon the advantages of adversarial learning paradigm, we present a two channel adversarial learning framework to learn spatio-temporal features by extracting ROI and its generated optical flow followed by a joint discriminator. We note that the introduction of person-ROI and difference loss function increases the performance. The major improvement in comparison to previous methods is the increase in AUC of PR curve. The optical flow information is also useful in the network and the fused method performs better than the raw thermal analysis only. In the future, we plan to extend the proposed techniques to detect falls using multiple camera modalities, including depth and IP cameras.

\small

\end{document}